\documentclass{article}

\usepackage{mylayout}
\usepackage[none]{hyphenat}
\usepackage{graphicx} 
\usepackage{amsmath} 
\usepackage{amssymb}  

\usepackage[document]{ragged2e}

\title{
Unsupervised Learning through Temporal~Smoothing and Entropy~Maximization
}

\author{Per Rutquist
\thanks{P. Rutquist is with the Institut für Mikrosystemtechnik,
        Albert-Ludwigs-Universität, Freiburg, Germany
        and with Tomlab Software AB, Sweden.
        {\tt\small per.rutquist@imtek.uni-freiburg.de}}%
}

\begin{document}

\maketitle

\begin{abstract}

This paper proposes a method for machine learning from unlabeled data in the form of a time-series. The mapping that is learned is shown to extract slowly evolving information that would be useful for control applications, while efficiently filtering out unwanted, higher-frequency noise.

The method consists of training a feedforward artificial neural network with backpropagation using two opposing objectives.

The first of these is to minimize the squared changes in activations between time steps of each unit in the network. This ``temporal smoothing''  has the effect of correlating inputs that occur close in time with outputs that are close in the L2-norm.

The second objective is to maximize the log determinant of the covariance matrix of activations in each layer of the network. This objective ensures that information from each layer is passed through to the next. This second objective acts as a balance to the first, which on its own would result in a network with all input weights equal to zero.

The method was tested in two experiments. In the first, a network learned from synthetic movies simulating a shaky camera pointed at the face of a ``clock''. The network reduced the input dimension from 784 quickly changing pixels to 16 slowly varying outputs. From these outputs the cosine and sine of the angle between the clock hands could be approximately recovered as affine functions of the outputs, even though they had not been shown to the network during training and were not recoverable as affine functions of the inputs directly. The network learned to extract relevant (to a control application) information, even though it had not explicitly been told what to consider as ``relevant.''

In a second experiment, learning from synthetic movies of moving hand-written digits from the MNIST dataset, a network found a representation such that nearest-neighbor classification achieves over 80 \% accuracy given only the first 18 labels in the training set.

\end{abstract}


\section{Introduction}

Machine learning has made it possible to use input sources such as video cameras in feedback control applications (e.g. self driving cars).  Artificial neural networks can be trained to extract the relevant information from images, reducing the input dimension. For example: a digital camera might provide several million pixel values per image, but control algorithms typically use no more than a few tens of inputs.

In supervised learning, each training sample $x_i$ (e.g. video frame) has an attached label $y_i$ (e.g. the car position on the road). A disadvantage of this approach is that the labels must be created by a human, and it can be an expensive and time-consuming process to create a sufficiently large dataset.

Humans are able to learn from far fewer labeled examples than are currently used in typical machine learning dataset.
For example, parents of a one-year old will note that their toddler can distinguish between a cat and a dog after having had between three and ten examples of each pointed out to him or her -- far from the 10 000 examples per category that are in the CIFAR-10 dataset commonly used to train computers on the same task.

If machine learning, like human learning, could use training sets of tens or hundreds -- rather than thousands or millions -- of labeled samples, then it would be possible to apply it to areas where it is currently not economical. In this paper we present a method for moving in this direction, using \emph{unsupervised} learning to find a map $x_i \rightarrow z_i$ such that, once this map is found, a small number of labeled pairs $(z_i, y_i)$ is sufficient for the machine to find a good approximation to the map from $z$ to $y$.

Humans do not start from nothing when learning to distinguish between cats and dogs. Infants spend countless hours looking at various objects, inspecting them from different angles and in different light. In artificial intelligence parlance, this would be called unsupervised learning, as the infants do not get any ``labels'' describing the object, angle and lighting conditions that they are observing. It is possible that this unsupervised learning teaches the infants how to perceive 3-D geometry and texture, which will later be useful in distinguishing between canines and felines.

A similar approach can be applied to machine learning. We consider a system where unlabeled data is ``free'', i.~e. available in unlimited quantity, in the form of a time series. (In the video camera example, this would consist of a very long movie.) The machine learning goal is to use this data to learn a mapping from the high-dimensional input space to a lower dimensional output space, such that relevant information is preserved while noise is discarded. This should be done without needing to tell the learning algorithm what we consider to be ``relevant''.

\subsection{Existing Approaches to Dimension Reduction}

 One of the most commonly used methods for dimension reduction is principal component analysis (PCA), where the low-dimensional output is a linear function of the input, consisting of the projection onto a subspace spanned by vectors corresponding to the largest singular values.
 
There are also nonlinear, unsupervised machine learning methods for dimension reduction. Examples include autoencoders and generative adversarial networks.

Autoencoders are neural networks that are trained to map input data to itself. By making one of the hidden layers smaller than the input dimension, the network is forced to find an efficient representation of the data. Learning can be facilitated by initializing weights using reduced Boltzmann machines~\cite{hinton2006reducing}.

A problem with autoencoders is how to define the loss function. For example, two images can be visually very similar and still differ substantially in every individual pixel value. Generative Adversarial Networks~\cite{goodfellow2014generative} solve this problem by training a separate ``discriminative'' network that tries to distinguish between data from the training set and the output of a ``generative'' network.

\subsection{Contribution and overview}

The unsupervised learning method presented in the following sections consists of balancing two opposing objectives: minimizing temporal correlation in activations on one hand, and maximizing information propagation on the other.

The first objective penalizes changes in activations over time. It is motivated by a ``slowness prior'' assumption; that relevant information is likely to be remain static or only vary slowly, while noise will vary more quickly. This is the same assumption that slow feature analysis \cite{wiskott2002slow} is based on. Penalizing temporal changes in activations as a regularization method has previously been investigated and found to have a similar effect as weight decay in the limiting case~\cite{stone1999temporal}.

For the second objective, the log determinant of the covariance matrix of activations in each layer is used as a measure of information propagation. For a multidimensional Gaussian distribution the differential entropy grows in proportion to this term. However, as pointed out by Salakhutdinov and Hinton~\cite{salakhutdinov2007learning}, the distribution learned by the network will not necessarily be Gaussian, and the network can cheat by creating ``hairball'' distributions, making the Gaussian approximation severely overestimate entropy.

Although both of these machine learning objectives have been investigated separately, to the author's knowledge they have not previously been used in combination as an unsupervised learning objective. The hypothesis, which seems to be confirmed by experiment, is that temporal smoothing will prevent the network from learning ``bad'' distributions, and at the same time the entropy objective will prevent the temporal smoothing from driving weights to zero.

In Section \ref{sec:obj}, the two machine learning objectives will be defined. Two different networks were trained using these opposing objectives. The first network was used to find information in a video feed, as detailed in Section \ref{sec:clock}, while the second, described in Section \ref{sec:mnist}, was trained to see planar images from different perspectives.

\section{Machine Learning Objective} \label{sec:obj}

We consider a feedforward network where the activation $a_{l,k,t}$ of unit $k$ in any layer $l$ at time step $t$ is computed as a function of the trainable parameters (weights and biases) of the layer and the activations $a_{l-1,\cdot,t}$ of the units in the preceding layer (or the inputs) at the same time step.

For the purpose of this discussion it does not matter if the network is dense, sparse, convolutional, or has some other structure, as long as it consists of layers and there is a well-defined way to compute gradients of activations with respect to the trainable parameters.

\subsection{Temporal Smoothing for Noise~Suppression} \label{sec:ts}

The temporal smoothing objective $f_\mathrm{TS}$ simply consists of a weighted sum of all the changes in unit activations, summed over all time-steps.

\begin{equation}
f_\mathrm{TS} = \sum_{t} \sum_{l} c_l \sum_{k} \left( a_{l,k,t} - a_{l,k,t-1} \right)^2
\end{equation}

Here $c_l$ are coefficients used to assign different weights to different layers, $l$ is the layer index and $k$ is the index of the unit within the layer.

Since this objective depends directly on activations in all layers of the network,
it is not susceptible to the problem of diminishing gradients that can occur in backpropagation through deep networks.

\subsection{Entropy Objective} \label{sec:e}

We will denote by $\mathrm{Cov}(X)$ the covariance matrix of a random column vector $X(t)$. It is equal to the expected value of the outer product of the vector with itself, after the mean has been removed

\begin{equation} \label{eq:ts}
  \mathrm{Cov}(X) = E\{(X - E\{X\})(X - E\{X\})^\top\}
\end{equation}

(In calculations, we will need to replace the expectancy by the mean over a batch of sample inputs, in order to keep computation time finite.)

We are interested in the activation covariance matrices in each layer of the network. Let $A_l$ denote the vector of activations in layer $l$, i.~e.

\begin{equation}
  A_l(t) = \left[ a_{l,1,t},\, \ldots,\, a_{l,N_l,t}\right]^\top .
\end{equation}

If the activation covariance matrix $\mathrm{Cov}(A_l)$ of a hidden layer $l$ does not have full rank, then there exists a network that produces exactly the same output using fewer units in layer $l$.
Such a network can easily be constructed since a covariance matrix that is singular implies that one of the elements of $A_l$ is always a linear combination of the other elements (plus a constant.) Hence, the unit corresponding to that element can be removed, and the remaining outgoing weights (and biases) modified to reproduce the effect of the removed unit.

A covariance matrix that is singular is thus an indication that information is not being propagated efficiently through the network. The determinant of such a covariance matrix is zero. Conversely, a higher value of the determinant of the covariance matrix (sometimes called the generalized variance) typically implies that more information is propagated. As already mentioned, the logarithm of this determinant grows in proportion to the differential entropy, if the distribution is Gaussian.

We therefore define an ``entropy'' objective $f_E$ as a weighted sum of the log determinant of the covariance matrices of each layer,
\begin{equation}
f_{\mathrm{E}} = \sum_l d_l \log \det \left(\mathrm{Cov}(A_l) \right)
\end{equation}
where the coefficients $d_l$ are tuning parameters that give varying importance to different layers.

Just like $f_\mathrm{TS}$, this objective is dependent on activations in all layers of the network, avoiding problems with diminishing gradients.

The gradient of a log determinant is easy to compute:
\begin{equation}
  \frac{\mathrm{d}}{\mathrm{d} \, \mathrm{vec}(M)} \log \det (M) = \mathrm{vec}(M^{-\top})
\end{equation}
where $M$ is an invertible matrix and $\mathrm{vec}$ denotes the operator that stacks the element of the matrix into a vector.

Assuming that the network is initialized properly, the covariance matrix will always be positive definite. (A singular covariance matrix would correspond to the maximization objective $f_\mathrm{E}$ going to minus infinity.) Hence, a Cholesky factorization can be used to compute both the determinant and the inverse needed for its gradient.

Since the effort of computing determinants (and their gradients) is independent of the batch size, it becomes insignificant for large batch sizes.

\subsection{The TS-E objective} \label{sec:ts-e}

The training objective that will be used in the subsequent sections simply consists of the difference between the two objectives described above.

\begin{equation} \label{eq:ts-e}
f_{\mathrm{TS-E}} = f_\mathrm{TS} - f_\mathrm{E}
\end{equation}

In the remainder of this text, we will refer to this as the TS-E (Temporal Smoothing minus Entropy) objective, and the neural networks that are trained by minimizing it will be referred to as TS-E networks.

\section{Example: ``Clock Hands''} \label{sec:clock}

In order to test the theory, two small-scale simulation studies were performed.

\begin{figure}
   \centering
   \includegraphics[scale=0.4]{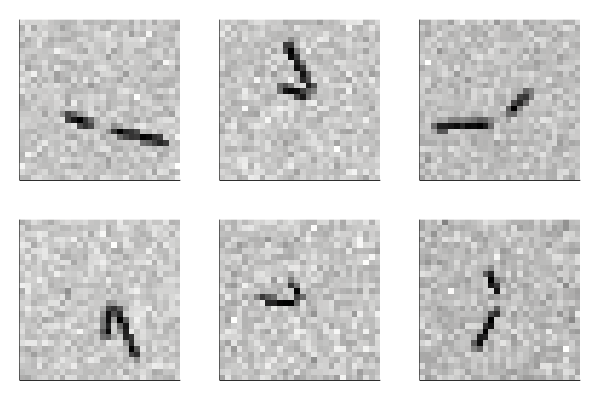}
   \caption{Samples of the simulated ``clock'' images. In addition to the pixel noise, there is also time-dependent ``camera shake'' in the position, scale and rotation angle of each image.}
   \label{fig:clock_samples}
\end{figure}

In the first simulation, low-resolution movies were produced simulating a
shaky, noisy low resolution camera pointed at the face of a clock.
The clock has two hands. In a steady frame of reference, the longer hand rotates continuously clockwise at a steady pace of one revolution per 400 frames, and the shorter hand rotates in the same direction at one fifth of that pace, so that the angle between the hands goes through one revolution every 500 frames in a cycle that repeats every 2 000 frames. However, the movie frames do not depict a steady frame of reference. The camera's 3D position as well as its rotation in the image plane follow a low-pass filtered random-walk. Example image frames can be seen in Figure~\ref{fig:clock_samples}.

\subsection{Training}

A feedforward network consisting of 784 ($28 \times 28$) inputs, eight fully connected hidden layers with 144, 121, 100, 81, 64, 49, 36 and 25 units respectively and a fully connected output layer containing 16 units was trained using the TS-E objective \eqref{eq:ts-e}. (Layer sizes were selected arbitrarily. Different layouts will be analyzed in future work.) The hyperbolic tangent ($\tanh$) function was used as activation function in all hidden layers and in the output layer.

The coefficients $c_l$ were chosen as $2^{l-1}$, i.~e. 1 for the first layer, 2 for the second, 4 for the third, and so on. The coefficients $d_l$ were all set to 10.

The TS-E network was trained using stochastic gradient descent with the ADAM algorithm~\cite{Kingma2014} within the Flux machine learning framework~\cite{innes:2018} in the Julia programming language~\cite{bezanson2017julia}. (No comparison between different solvers or hyperparameters was performed.)

\subsection{Results}

Once unsupervised learning had finished, the network was tested using a new, previously unseen movie generated using the same stochastic model as had been used to create the training data.

\begin{figure}
   \centering
   \includegraphics[scale=0.6]{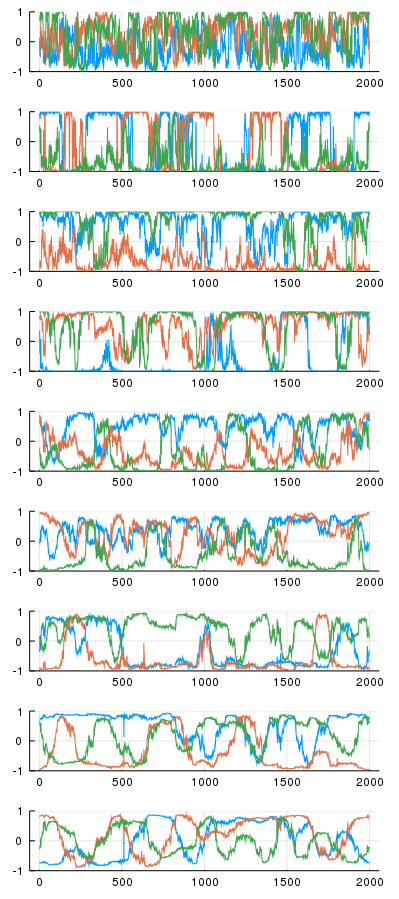}
   \caption{Unit activations as function of time for three randomly selected units in each layer. As we progress from the first layer (top) to the last (bottom) the volatility of the signal can be seen to diminish.}
   \label{fig:clock_activations_vs_time}
\end{figure}

Figure~\ref{fig:clock_activations_vs_time} shows a few randomly selected activations for each layer. It can clearly be seen that activations fluctuate more slowly in each subsequent hidden layer compared to the preceding one. Although the input is a time series, it should be pointed out again that the network itself contains no dynamic state. The only reason why each layer shows less fluctuations than the preceding one is that the learning algorithm has found ``features'' (combinations of activations) in the upstream layer that vary more slowly than the individual activations themselves.

In order to investigate whether the network retains useful information, we attempted to re-construct the angle $\alpha$ between the shorter and the longer of the clock hands from the output of the network. This angle is one of the features that a human observer would see as most prominent when looking at the images in Figure~\ref{fig:clock_samples} so we would like for this information to be still present in the output.

The angle $\alpha$ can only be measured modulo whole rotations, and is therefore not a continuous function of time. However, its sine and cosine do vary continuously. Hence, it was attempted to recover $\sin \alpha$ and $\cos \alpha$ from the network output using linear regression. A set of 100 randomly chosen (essentially uncorrelated) frames from a movie sequence were labeled with the ground truth $(\cos \alpha,\, \sin \alpha)$ value, and a linear least squares problem was solved to estimate these as affine functions of the 16 outputs of the TS-E network.

\begin{figure}
   \centering
   \includegraphics[scale=0.4]{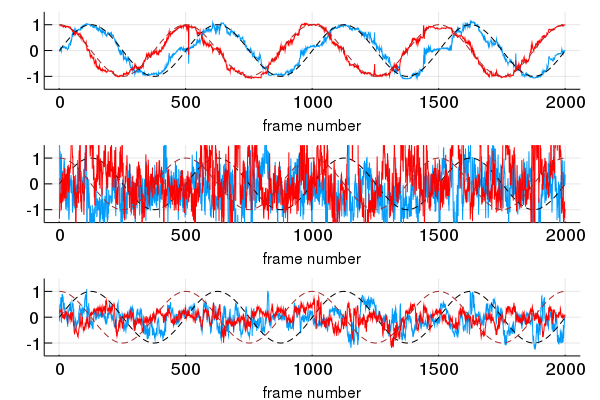}
   \caption{The estimated cosine and sine of the intra-hand angle $\alpha$ as affine functions of the output of the TS-E network (top), of downsampled pixels (middle), and of the principal components (bottom). Dashed lines show the cosine and sine of the ground truth $\alpha$ used to create the simulated movie frame.
   The input data consisted of a movie that the network had never seen during training, but which was generated using the same stochastic model.} \label{fig:sin_100}
   
\end{figure}

The uppermost graph in Figure~\ref{fig:sin_100} shows the time series as estimated by the above algorithm. The plotted curve shows the raw estimates, computed independently for each component in each frame.
Below it, for comparison, the results of linear regressions in subspaces of the raw inputs, either using principal component analysis or linear downsampling (taking the means of $4 \times 4$ pixel clusters) are shown.

In the top graph, the estimates computed from the network output are visibly close to the ground truth reference curves, with a root-mean-square (rms) error of approximately 0.2.
On the other hand, the estimates computed from linear combinations of the inputs (principal components or downsampling) were completely useless, yielding signals that were essentially uncorrelated to the ground truth. (Using more labeled data for the linear regression did not help.)
Not only has the TS-E network separated data from noise; the nonlinear transformation of the network has revealed information that was not available as a linear function of the raw pixel data!

\section{Example: MNIST Handwritten Digits} \label{sec:mnist}

The MNIST handwritten digits dataset~\cite{lecun-mnisthandwrittendigit-2010} is one of the most frequently used datasets in machine learning.
It consists of a training set of 60 000 images and a test set of 10 000 images, each labeled with categories $\in 0 \ldots 9$

The goal of this experiment was not to achieve the best possible accuracy, but to investigate whether the nonlinear transformation of a TS-E network could be beneficial, i.e. simplify the subsequent task of classifying the images.

\begin{figure}
   \centering
   \includegraphics[scale=0.4]{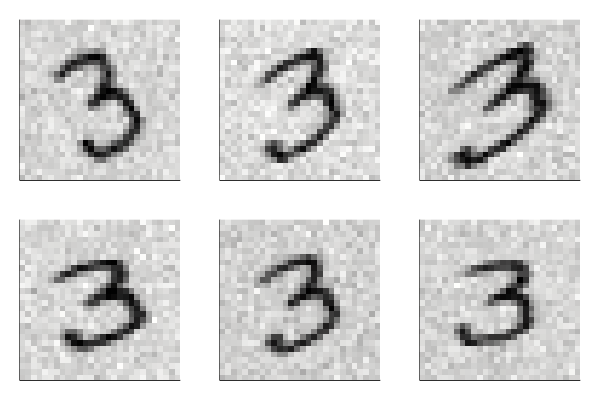}
   \caption{Sample frames from a synthetic movie of one image from the MNIST dataset.}
   \label{fig:mnistframes}
\end{figure}

The dataset itself does not contain any time-series data, but movies were synthesized of a simulated shaky, noisy camera pointed at the images, affecting the position, scale, rotation, shear and aspect ratio, similar to how movies were generated in Section \ref{sec:clock}. (More advanced image-distortion techniques have been used by others for augmentation of this dataset, see for example~\cite{wong2016understanding}, but were not investigated in this work due to time constraints.)
Of the images in the training set, half (i.e. 30 000 images) were used for synthesizing movies (their labels thrown away). The movies were generated using a low-pass filtered random-walk model for the camera parameters, reverting to a mean position (corresponding to the undistorted image) with a time constant of 24 frames. The variance parameters of the stochastic walk were selected by visually inspecting the resulting movies and ensuring that the digits were not noticeably more difficult to identify after adding the distortion and noise. A few sample frames can be seen in Figure~\ref{fig:mnistframes}.

\subsection{Training}

Unsupervised training was performed in the same manner as described in Section~\ref{sec:clock}. The training input consisted of 240-frame movies generated from each training image. Movies were generated anew on each training iteration, so that the optimization algorithm would not overfit to the random camera movement.
Because a separate movie clip was generated for each image, the temporal smoothing algorithm did not see any transitions between digits.
Each batch consisted of 1000 movie clips, so each covariance matrix was computed from 240 000 frames based on 1000 different images.

\subsection{Results}

In order to evaluate the effectiveness of the transformation defined by the network, it was investigated how well nearest-neighbor classification works in the space of the network output. Nearest-neighbor classification simply consists of finding the training sample which has the shortest distance to a given test sample, and assigning the class of that training sample to the test sample. The only tunable parameter is the choice of distance measure, for which we used the Euclidean distance in the outputs of the TS-E network. Sets of various sizes from 18 to 30 000 images that had not been used for movie generation were used as training sets, and the accuracy was evaluated on the 10 000 images of the test set. For comparison, the accuracy using Euclidean distances in the principal components, in downsampled pixels, and in the raw input pixels was also computed.

\begin{figure}
   \centering
   \includegraphics[scale=0.4]{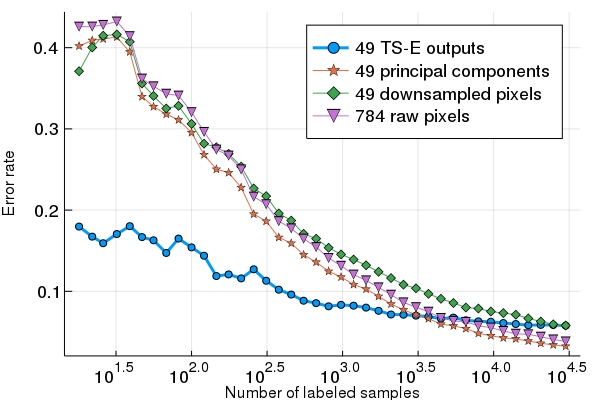}
   \caption{Error rate of nearest-neighbour classification on the MNIST dataset, as function of the number of labeled samples. In the output space of the TS-E network, the nearest labeled neighbour is of the right class over 80 \% of the time, even with only 18 labeled samples, which is the first point in the test set where all digits have appeared at least once. The TS-E network outperforms linear methods until the number of labeled example reaches the thousands.}
   \label{fig:mnist_accuracy}
\end{figure}

Figure~\ref{fig:mnist_accuracy} shows accuracy as a function of the number of training samples. The smallest number of labeled examples used was 18. This is the first point in the chosen dataset where each digit has occurred at least once. At this point, nearest neighbor classification in the TS-E output had an error rate of less than 20 percent, approximately half that of either of the other three distance measures. Clearly the nonlinear transformation learned by the network has brought similarly classified images together, even though no class information was used in training.

The network output continues to give the highest accuracy as the number of labeled examples goes to 1000, but at 10 000 or more labeled examples, distances in principal components or raw pixels perform better.


\section{Discussion}

\begin{figure}
   \centering
   \includegraphics[scale=0.4]{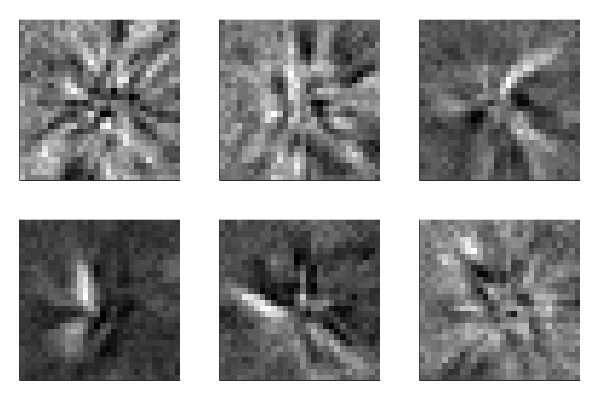}
   \caption{The input weights of six randomly selected units in the first hidden layer of the TS-E network used to process the ``clock hands'' in section \ref{sec:clock}.
   }
   \label{fig:clock_first_layer_weigths}
\end{figure}

Convolutional neural networks were not used in the case studies. There was no information about the 2D layout of the 784 pixels encoded in each frame. the TS-E network seems to have recovered this information on its own, and looking at the weights in the first layer, there is clearly some 2D structure. (See Figure \ref{fig:clock_first_layer_weigths}.) It seems very likely that the method would benefit from using convolutional networks, and this would make it possible to work with larger frame sizes in the movies used for training.

During the experiments, new random inputs were generated throughout the unsupervised training phase.
This corresponds to an assumption that the amount of training data exceeds what can be utilized. The justification for this assumption is that the training data simply consists of raw data from the system, without any labels or need for manual processing.

Movie generation (which ran on the CPU) was slower than the gradient evaluations (which ran on the GPU) by an order of magnitude.
Due to time constraints, no effort was made to optimize this part of the code or port it to the GPU, and therefore it was not possible to make proper timing measurements of the unsupervised learning. During each of the experiments, training ran for approximately three hours on a single personal computer after which it was halted for practical reasons even though longer runs may have yielded better convergence.

In the examples, the TS-E networks were used purely as pre-processing steps to linear regression or a nearest neighbor classifier. It would be possible (and likely useful) to instead continue the network with more layers for the final processing, training the entire network with back-propagation in supervised learning. The unsupervised learning could then be seen as an initialization step for supervised learning, similar to how reduced Boltzmann machines are be used in~\cite{hinton2006reducing}.

Yet another option is to not estimate the physical state of the system at all, but instead use the output of the TS-E network as input to a model-free algorithm such as Q-learning.

\section{Conclusion and Future Work}

Unsupervised learning from time series data can be accomplished using a combination of a temporal smoothing and a data propagation (entropy) objective. In simulation, the method was able to find information that a human would consider relevant; the relative orientation of objects in the frame.
This was achieved even though it required a nonlinear transformation to be learned and without the network having access to examples of the sought information during training.

In a second simulation, a network learned from synthetic movies to map images into a representation such that the same image seen from different perspectives yields outputs that are close to each other in Euclidean distance. For smaller datasets of less than 1000 labeled samples, this drastically improved the accuracy of nearest-neighbor classification compared to linear mappings.

Future work will extend the method to convolutional networks, and use more realistic input data and more specialized optimization methods. It is also planned to replace the entropy maximization objective by lower bounds on entropy in each layer, using a solver that can incorporate such constraints.



\begin{thebibliography}{10}

\bibitem{hinton2006reducing}
Geoffrey~E Hinton and Ruslan~R Salakhutdinov.
\newblock Reducing the dimensionality of data with neural networks.
\newblock {\em science}, 313(5786):504--507, 2006.

\bibitem{goodfellow2014generative}
Ian Goodfellow, Jean Pouget-Abadie, Mehdi Mirza, Bing Xu, David Warde-Farley,
  Sherjil Ozair, Aaron Courville, and Yoshua Bengio.
\newblock Generative adversarial nets.
\newblock In {\em Advances in neural information processing systems}, pages
  2672--2680, 2014.

\bibitem{wiskott2002slow}
Laurenz Wiskott and Terrence~J Sejnowski.
\newblock Slow feature analysis: Unsupervised learning of invariances.
\newblock {\em Neural computation}, 14(4):715--770, 2002.

\bibitem{stone1999temporal}
James~V Stone and Nicol Harper.
\newblock Temporal constraints on visual learning: a computational model.
\newblock {\em Perception}, 28(9):1089--1104, 1999.

\bibitem{salakhutdinov2007learning}
Ruslan Salakhutdinov and Geoff Hinton.
\newblock Learning a nonlinear embedding by preserving class neighbourhood
  structure.
\newblock In {\em Artificial Intelligence and Statistics}, pages 412--419,
  2007.

\bibitem{Kingma2014}
Diederik~P. Kingma and Jimmy Ba.
\newblock Adam: {A} method for stochastic optimization.
\newblock {\em CoRR}, abs/1412.6980, 2014.

\bibitem{innes:2018}
Mike Innes.
\newblock Flux: Elegant machine learning with julia.
\newblock {\em Journal of Open Source Software}, 2018.

\bibitem{bezanson2017julia}
Jeff Bezanson, Alan Edelman, Stefan Karpinski, and Viral~B Shah.
\newblock Julia: A fresh approach to numerical computing.
\newblock {\em SIAM review}, 59(1):65--98, 2017.

\bibitem{lecun-mnisthandwrittendigit-2010}
Yann LeCun and Corinna Cortes.
\newblock {MNIST} handwritten digit database.
\newblock 2010.

\bibitem{wong2016understanding}
Sebastien~C Wong, Adam Gatt, Victor Stamatescu, and Mark~D McDonnell.
\newblock Understanding data augmentation for classification: when to warp?
\newblock In {\em 2016 international conference on digital image computing:
  techniques and applications (DICTA)}, pages 1--6. IEEE, 2016.

\end{thebibliography}
\end{document}